\pdfoutput=1
\documentclass[11pt]{article}

\usepackage{ACL2023}

\usepackage{times}
\usepackage{latexsym}

\usepackage[T1]{fontenc}
\usepackage[utf8]{inputenc}

\usepackage{microtype}

\usepackage{inconsolata}

\newcommand{\thickhline}{\noalign{\hrule height 1pt}}

\usepackage{hyperref}       % hyperlinks
\usepackage{url}            % simple URL typesetting
\usepackage{booktabs}       % professional-quality tables
\usepackage{nicefrac}       % compact symbols for 1/2, etc.
\usepackage{xcolor}         % colors
\usepackage{multirow}
\usepackage{bm}

\usepackage{times}
\usepackage{latexsym}

\usepackage{amsmath}

\usepackage{algorithm}
\usepackage{algpseudocode}
\usepackage{graphicx}  %Required
\usepackage{CJK}
\usepackage{bbm}
\usepackage{float}
\usepackage{xcolor,colortbl}

\usepackage{bigstrut,bigdelim}
\usepackage{paralist}
\usepackage{diagbox}
\usepackage{wrapfig}
\usepackage{verbatim}
\usepackage{enumitem}

\usepackage{subcaption}
\usepackage{multicol}
\usepackage{multirow}
\usepackage{amssymb}
\usepackage{xcolor}
\usepackage{caption}
\usepackage{cleveref}
\usepackage{acronym}
\usepackage{xspace}
\usepackage{pifont}

\makeatletter
\DeclareRobustCommand\onedot{\futurelet\@let@token\@onedot}
\def\@onedot{\ifx\@let@token.\else.\null\fi\xspace}

\def\ie{\emph{i.e}\onedot}

\def\vs{\emph{vs}\onedot}

\makeatother

\DeclareMathOperator*{\argmax}{arg\,max}

\newcommand{\cmark}{\ding{51}}%
\newcommand{\xmark}{\ding{55}}%

\acrodef{nlp}[NLP]{natural language processing}
\acrodef{vqa}[VQA]{Visual Question Answering}
\acrodef{cs}[CS]{coherence scoring}
\acrodef{avsd}[AVSD]{coherence scoring}
\acrodef{roi}[RoI]{Region-of-Interest}

\title{VSTAR: A Video-grounded Dialogue Dataset for Situated Semantic Understanding with Scene and Topic Transitions}

  \author{Yuxuan Wang$^{1,2,3}$\Thanks{~This work was partially conducted when Yuxuan Wang was a research intern at BIGAI.} , Zilong Zheng$^{2}$\Thanks{~Correspondence to Zilong Zheng and Dongyan Zhao.} , Xueliang Zhao$^{1}$ \\ \textbf{Jinpeng Li$^{1}$, Yueqian Wang$^{1}$, Dongyan Zhao$^{1,2,3,4\dagger}$} \\
$^1$ Wangxuan Institute of Computer Technology, Peking University, Beijing, China \\
$^2$ Beijing Institute for General Artificial Intelligence (BIGAI), Beijing, China \\
$^3$ Center for Data Science, AAIS, Peking University, Beijing, China \\
$^4$ National Key Laboratory of General Artificial Intelligence, Beijing, China \\
\texttt{ \{wyx,lijinpeng\}@stu.pku.edu.cn, zlzheng@bigai.ai} \\ 
\texttt{\{xl.zhao,wangyueqian,zhaody\}@pku.edu.cn} \\ 
\url{https://vstar-benchmark.github.io/}
}

\makeatletter
\let\@oldmaketitle\@maketitle% Store \@maketitle
\renewcommand{\@maketitle}{\@oldmaketitle% Update \@maketitle to insert...                    
\vspace*{.4in}
  \includegraphics[width=\linewidth]{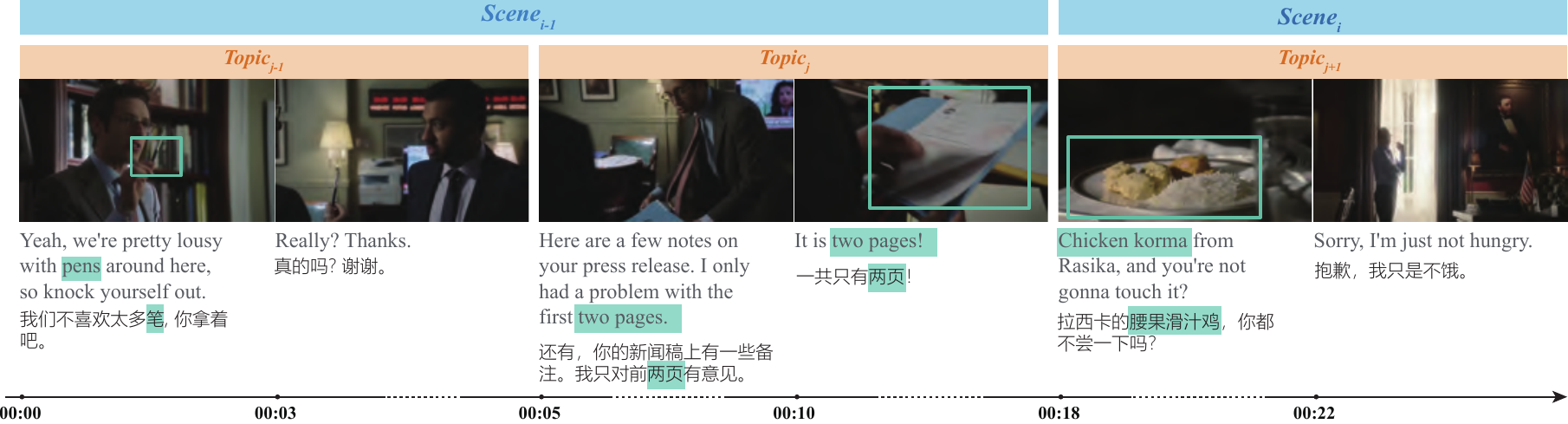}
   \captionof{figure}{\textbf{An overview of our VSTAR dataset.}  A 30 seconds video clip involves two dialogue scenes in which the environments and interlocutors are totally different.}\label{fig:intro}\bigskip
  }% ... an image
\makeatother

\begin{document}
% \twocolumn[{
% \renewcommand\twocolumn[1][]{#1}
% \maketitle
% \begin{center}
%     \captionsetup{type=figure}
%     \includegraphics[width=1\textwidth]{figs/STVD_example2.pdf}
%    \captionof{figure}{\textbf{An overview of our VSTAR dataset.}  A 30 seconds video clip involves two dialogue scenes in which the environments and interlocutors are totally different.}\label{fig:intro}
% \end{center}
% }]
\maketitle
% \twocolumn[{
% \renewcommand\twocolumn[1][]{#1}
% % \maketitle
% \maketitle

% \begin{center}
%     \captionsetup{type=figure}
%     \includegraphics[width=1\textwidth]{figs/STVD_example2.pdf}
%    \captionof{figure}{\textbf{An overview of our VSTAR dataset.}  A 30 seconds video clip involves two dialogue scenes in which the environments and interlocutors are totally different.}\label{fig:intro}
% \end{center}
% }]

\begin{abstract}
% video-grounded dialogue benchmark dataset 
% Recent years have witnessed a strongly increasing interest in video-grounded dialogue tasks. \wyx{Existing benchmarks concentrate on annotations linking visual cues and semantics. In real word, it is almost impossible to infer what a person talk from his behavior. But we do talk differently in different occasions.}
Video-grounded dialogue understanding is a challenging problem that requires machine to perceive, parse and reason over situated semantics extracted from weakly aligned video and dialogues. 
Most existing benchmarks treat both modalities the same as a frame-independent visual understanding task, while neglecting the intrinsic attributes in multimodal dialogues, such as scene and topic transitions.   
In this paper, we present \textbf{\underline{V}ideo-grounded \underline{S}cene\&\underline{T}opic \underline{A}wa\underline{R}e dialogue~(VSTAR)} dataset, a large scale video-grounded dialogue understanding dataset based on 395 TV series. 
Based on VSTAR, we propose two benchmarks for video-grounded dialogue understanding: scene segmentation and topic segmentation, and one benchmark for video-grounded dialogue generation. Comprehensive experiments are performed on these benchmarks to demonstrate the importance of multimodal information and segments in video-grounded dialogue understanding and generation.
% \footnote{All data and baseline code will be publicly available.}
\end{abstract}

\section{Introduction}

% \zzl{Visual Dialogue} 

\begin{quote}
    ``Every film should have its own world, a logic and feel to it that expands beyond the exact image that the audience is seeing.''
    \quad{}\quad{}\quad{}\quad{}\quad{}--- Christopher Nolan
\end{quote}

Have you seen movie ``Memento''? If so, you may feel astonished for the mysterious movie plots but also clap for the excellent directing and editing skills of director Christopher Nolan. In this movie, the main character plays a role of a murder victim with short-term memory. Truth, memories, fake stories are cut into pieces and re-ordered in the movie.
Such nonlinear narrative structure is not unique to Nolan's movie but is frequently used in story-based videos to increase the mysteriousness. The audience has to identify the type and chronological order of each scene in order to understand the entire movie. We call such ability as \textbf{scene transition identification}. 
In our daily dialogue, similarly, topic shifting commonly occurs during chit-chatting. One typical example is a speaking trick ``callback'' that is used by comedians, where a previously mentioned joke verbatim is repeated later.
The listener has to comprehend relations between different topics in order to understand the entire discourse. We call such ability as \textbf{topic transition identification}. 

However, these abilities have barely been investigated in modern literature of multimodal conversational understanding. Over the recent years, we have witnessed a growing trend of modeling longer and more diverse video-grounded dialogues due to growing computing capacity and hardware acceleration. However, most tasks are based on multi-turn \ac{vqa} for its simplicity in answer evaluation~\cite{Das2017VisualD,AlAmri2019AudioVS}, which yields huge differences compared with realistic conversations.  

% Visual Dialogue / Video Dialogue
% Building a dialogue agent that can comprehend multi-modal information has long been the goal of general artificial intelligence. Recently, the dialogue system involving visual and language information has attracted attention of many researchers with a amount of datasets with novel benchmarks being proposed. Visual Dialogue~\cite{Das2017VisualD, Seo2017VisualRR, Vries2017GuessWhatVO, Chattopadhyay2017EvaluatingVC} treated the problem as a multi-turn \ac{vqa}~\cite{Agrawal2015VQAVQ, Gao2015AreYT, Malinowski2014AMA, Zhu2016Visual7WGQ} that aims to develop dialogue agent that can answer question by given a dialogue history and its corresponding image. 
% Recently visual dialogue have been extended to video domain~\cite{AlAmri2019AudioVS, Le2021DVDAD}. In this problem, questions about a given video are positioned in a multi-turn QA-pairs. In each QA-pair turn, a question usually exhibits different types of cross-turn relations to other questions in prior turns. To mimic the situations where common dialogue happen,~\cite{Pasunuru2018GameBasedVD, Meng2020OpenViDialAL, Wang2021OpenViDial2A} construct open-domain video-grounded dialogue datasets. Due to the difficulty of obtaining daily conversation in real life, These datasets are constructed from live-broadcast, Movies and TV Series. 

\begin{table*}[t!]
\centering
\caption{\textbf{Comparisons of different multimodal dialogue datasets.} }
\label{tab:VSTAR}
\resizebox{\linewidth}{!}{
\begin{tabular}{lcccccccc}
\thickhline
            \textbf{Dataset} & \textbf{Vision} & \textbf{Language} & \textbf{Scene} & \textbf{Topic} & \textbf{\# Dialogues}   & \textbf{\# Turns}  &  \textbf{Turns/Clip} &  \textbf{Words/Turn}   \\
            \hline
            VisualDialog & Image  & QA  &  \xmark & \xmark & 120K &  2.4M & 20.0 & 4.0 \\
            Twitch-FIFA & Live Video & Dialogue  &  \xmark & \xmark   & 15K   & 161K   & 10.4  & 6.0 \\
            AVSD & Recorded Video  & Dialogue  &  \xmark & \xmark    & 11K   & 118K   & 20.0  & 9.5  \\
            MoiveNet$^\dagger$ & Movies & Dialogue  &  \cmark & \xmark & - & 421K & - & 7.2 \\
            OpenViDial 2.0 & Movies \& TV Series & Dialogue  &  \xmark & \xmark   & 116K  & 5.6M  & 48.0 & 8.3 \\ \hline
            VSTAR~(Ours) & TV Series  & Dialogue   &  \cmark & \cmark & 185K      & 4.6M      & 25.1  & 6.7 \\ 

% VSTAR~(Ours) & Video   & 185K      & 4.8M      & 25.1  & 6.7 & 1.4 & 2.7  \\ 
% \hline
\thickhline
\multicolumn{9}{l}{$\dagger$: We compute statistics of the sub-dataset with scene boundary annotations.}
\end{tabular}
}

\end{table*}

To fill the gap between the current multimodal dialogue systems and realistic video-grounded dialogue understanding, in this work, we introduce a new challenge, \textbf{\underline{V}ideo-grounded \underline{S}cene\&\underline{T}opic \underline{A}wa\underline{R}e dialogue~(VSTAR)} dataset, aiming to address the frequent scene and topic transition within open-domain video-grounded dialogues. We start by constructing a VSTAR dataset by collecting a total of 395 TV series and 8,159 TV episodes together with their corresponding story-line and keywords; \Cref{fig:intro} shows a typical example. To test the machine's abilities on scene and topic transition identifications, we annotate video scene boundary and dialogue topic boundary through determining the semantic transition among short video-dialogue turn pairs. \Cref{tab:VSTAR} shows the statistics and main differences of our new dataset compared with previous multimodal dialogue benchmark. 
Below, we point out the distinguished challenges existing in VSTAR:

\textit{Complicated video understanding.} We carefully selected story-based TV series as the main data source. Compared with movies or homemade short videos, TV series containing more complicated plots: there are many remarkable scene transitions and topic shifts within each video clip, which brings in extra challenge in video understanding. 

\textit{Multimodal scene\&topic transition identification.} It is worth noting that identifying scene and topic boundary in a video clip is non-trivial. Besides the complexity of long videos, both identifications require reasoning over multi-modalities. For example in \Cref{fig:intro}, it is hard to separate topics or scenes solely based on visual cues or dialogue text. 
% A scene can be viewed as a sequence of shots sharing semantic coherence. Thus, it is not easy to find the scene transition only by the visual cues. Naturally, the corresponding conversations in the videos can be treated as effective auxiliary information in the task. 
% Similarly, video is helpful in discovering the dialogue topic boundary.

\textit{High-level contextual information.}  In VSTAR, the object-level links between video and language are much  weaker than high-level semantic connections.  Different from showing captions that are directly related with videos as \ac{vqa}-based tasks do, reasoning over contents in VSTAR requires the capability of high-level multimodal contextual understanding, \ie, making connections between scenes, topics, and multimodal contexts.
% In summary, finding scene/topic transition is vital in open-domain video-grounded dialogue understanding. 

% There are totally 395 TV series with 8159 TV episodes and 4.6M utterances in the VSTAR. Each TV episodes is provided it corresponding meta data including story-line and keywords. Compared with movies, there is less monologue and artistic expression in the TV series. We annotate video scene boundary and dialogue topic boundary with the help of multi-modal information to support video-grounded dialogue understanding. Specifically, the annotations in VSTAR include 265K scene boundaries and 499K dialogue topic boundaries. 
% To the best of our knowledge, VSTAR is the first video-grounded dialogue dataset annotated with scene and topic boundaries.

% 为什么要有多模态切换：1）视频中本来就有非常多的场景切换，要理解上下文需要搞清楚这些切换，防止噪声，文本同理 2）单一模态可能在浅层特征上有比较高的相似度，所以需要不同模态的信息帮助理解区分场景/话题信息。3) 因为视频文本不是直接相关的，不是一一对应的，所以整个场景提供的 high-level 信息比较重要
% Compared with homemade short videos, the story-based videos, such as movies and TV series, contain more complicated plots. Specifically, there are many remarkable changes of scenes, which are traditionally viewed as basic semantic units. For example,  
% In a scene, the discourse-level topic shifts are common phenomenons. Therefore, figuring out the scene boundaries in video and topic boundaries in dialogue are important for long video understanding. 

% \zzl{need a paragraph for the importance and difference of VSTAR. For example, scene segmentation and topic segmentation}
We benchmark our dataset via three challenging tasks: scene segmentation, topic segmentation, and response generation. Detailed task formulation and evaluation are introduced in \Cref{sec:benchmark}.  Moreover, we propose a sliding-window based discriminative model (SWST) for segmentation tasks and an autoregressive  generative model (AVDT) for dialogue generation. Extensive experiments are performed to evaluate our model and baseline methods in video-grounded dialogue understanding and generation; refer to \Cref{sec:exp} for detailed results and analysis. 

% Based on the data and annotations in VSTAR, we set up three challenging benchmarks for video-grounded dialogue understanding and generation: (1) video-grounded dialogue scene segmentation, (2) video-grounded dialogue topic segmentation, (3) multi-modal dialogue generation. Concretely, given a video clip with dialogues, the segmentation tasks is to find all the semantic transitions in videos and dialogues, while the generation task require to generate a response according to the given multi-modal context. We represent examples in Fig~\ref{fig:intro}. A 30 seconds video a clip involve two dialogue scenes in which the environments and interlocutors are totally different. In the first scene there are two players first talking about pen then press release. Although they are in the same place, their dialogues share little connection as time goes by. As a result, the video and dialogue in $topic_{j}$ are enough for generating a consistent response of the query "I only had a problem with the first two pages". Such phenomenons are common in story-based videos. Based on abundant annotations in the VSTAR, we propose a sliding-window based discriminative model for segmenting task and an autoregressive  generative model for generation task. Extensive experiments are executed to present the advantages of our annotations in video-grounded dialogue understanding and generation. 
% Movie -> TV Series
% \zzl{One paragraph for benchmark, evaluation, metric}

In summary, our contributions are three-fold: (i) we collect and annotate VSTAR, a large-scale video-grounded dialogue dataset with scene and topic boundary annotations; (ii) we formalize three challenging tasks regarding video-grounded scene and topic segmentation and dialogue generation; (iii) we benchmark three tasks and analyze experimental results with baseline methods and two new transformer-based models: SWST and AVDT.
% \begin{itemize}
    % \item We introduce three novel tasks for video-grounded dialogue understanding and generaiton:(1) video-grounded dialogue scene segmentation, (2) video-grounded dialogue topic segmentation, (3) multi-modal dialogue generation. 
    % \item We collect VSTAR, a large-scale video-grounded dialogue dataset with dialogue scene and topic boundary annotations. 
    % To the best of our knowledge, this is the first video-grounded dialogue dataset that combines multi-modal information to find the semantic transition in video and dialogue.
    % \item We benchmark our dataset with a transformer-based architecture.
% \end{itemize}

% \zzl{Contribution Summary}

\section{Related work}
% \zzl{Too long for each subsection}
% \paragraph{Multimodal Dialogue Dataset}
Recently, multimodal dialogue systems have attracted interest of many researchers with a number of benchmarks and datasets proposed. 
VisualDialog~\cite{Das2017VisualD, Seo2017VisualRR, Vries2017GuessWhatVO,Chattopadhyay2017EvaluatingVC,zheng2019reasoning} treated the problem as a multi-turn \ac{vqa}
% ~\cite{Agrawal2015VQAVQ, Gao2015AreYT, Malinowski2014AMA, Zhu2016Visual7WGQ} 
that aims to develop a dialogue agent that can answer questions given dialogue histories and corresponding images. 
Empowered by today's high computational capacity, similar dialogue systems have been extended to the video domain~\cite{AlAmri2019AudioVS, Le2021DVDAD}: questions about a given video are positioned in a multi-turn QA-pairs. In each QA-pair turn, a question usually exhibits different types of cross-turn relations to other questions in prior turns. To mimic the situations where common dialogues happen. Twitch-FIFA~\cite{Pasunuru2018GameBasedVD} introduces a video-context dialogue dataset based on live-broadcast soccer game videos and chats from \textit{Twitch.tv}. \citet{Meng2020OpenViDialAL,Wang2021OpenViDial2A} construct OpenViDial dataset based on movies and TV series. However, the subtitles extracted by OCR in this dataset contain much noise and monologue. 
One of the closest work to us is the recently proposed MovieNet~\cite{Huang2020MovieNetAH}, a large scale dataset for movie understanding. 
% This dataset consists of 1,100 movies and 318 of them are annotated with scene labels. 
% Benifit from this large-scale labeled training data,~\cite{Chen2021ShotCS,Mun2022BoundaryawareSL} achieved considerable improvements to previous methods. 
In this work, shot is viewed as the minimal visual unit of a movie. However, shot boundary is annotated automatically by existing tool~\cite{Sidiropoulos2011TemporalVS}, which leads to inaccurate predictions. Besides, few methods take the multi-modal information as input which make current models learn little high-level semantic information.

\section{The VSTAR Dataset}

\begin{figure*}
    \centering
    \begin{subfigure}[b]{0.70\linewidth}
    \includegraphics[width=\linewidth]{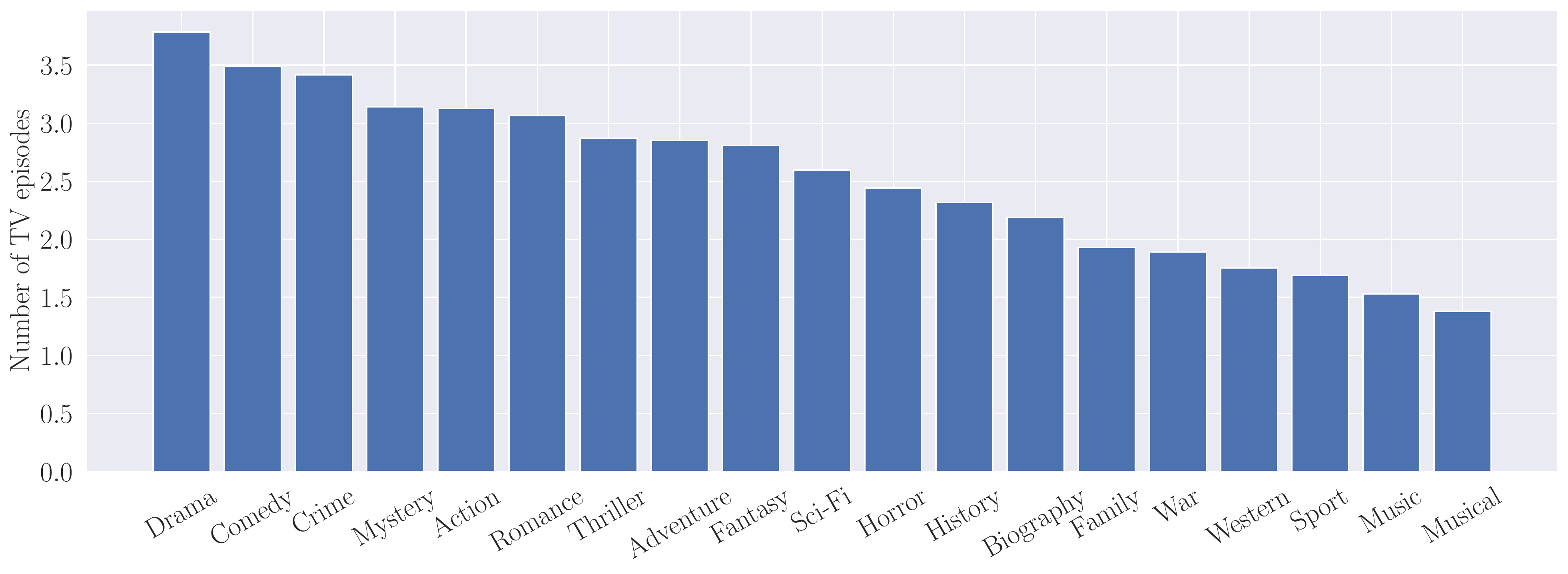}
    \caption{Distribution of VSTAR Genres (y-axis in log scale).}
    \label{fig:genres}
    \end{subfigure}
    \hfill
    \begin{subfigure}[b]{0.28\linewidth}
        \centering
        \includegraphics[width=\linewidth]{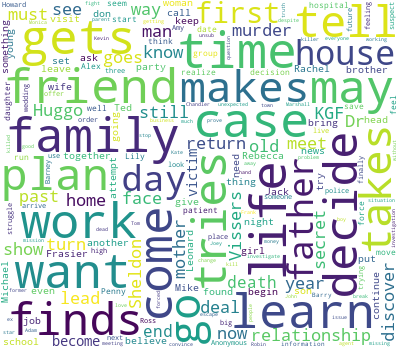}
        \caption{Story-lines wordcloud}
        \label{fig:wordcloud}
     \end{subfigure}
    \caption{Statistics of metadata in VSTAR}
    \label{fig:metadata}
\end{figure*}

\begin{figure*}
    \centering
    \begin{subfigure}[b]{0.49\linewidth}
    \centering
        \includegraphics[width=\linewidth]{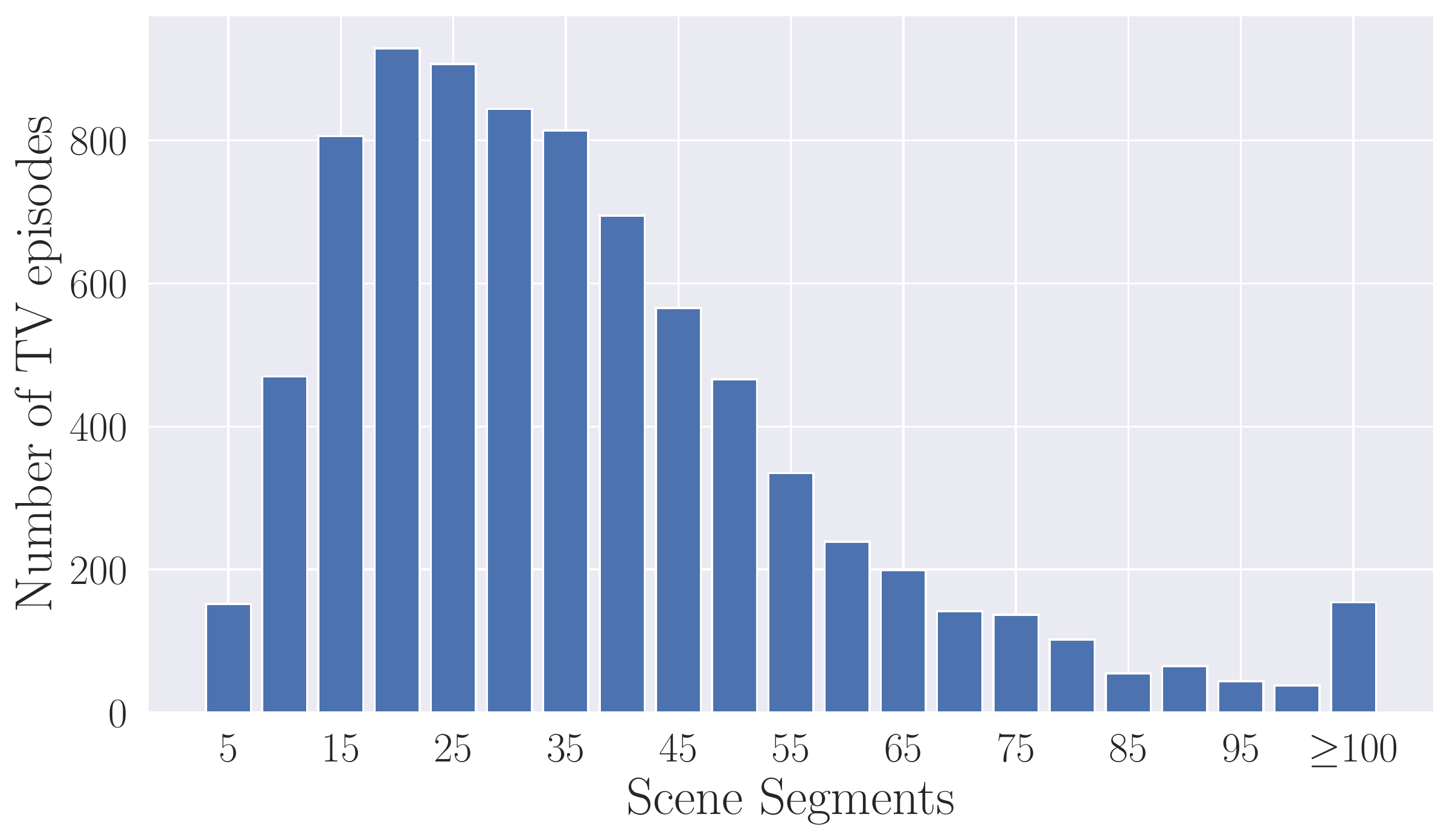}
        \caption{Distribution of \# scene segments}
        \label{fig:session_segments}
    \end{subfigure}
    \hfill
    \begin{subfigure}[b]{0.49\linewidth}
    \centering
        \includegraphics[width=\linewidth]{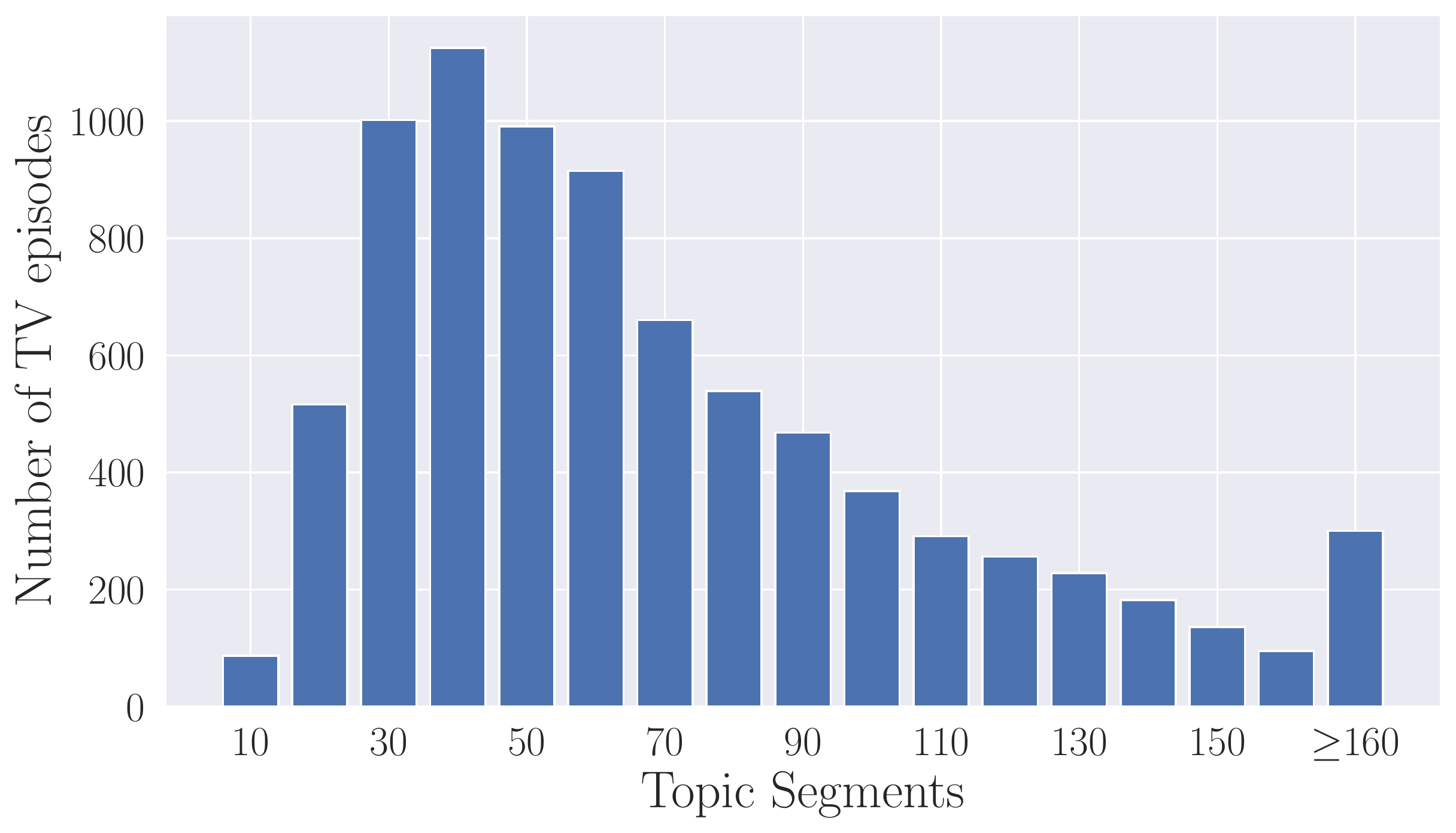}
        \caption{Distribution of \# topic segments}
        \label{fig:topic_segments}
    \end{subfigure}
    \caption{Statistics of annotations in VSTAR}
    \label{fig:annotations}
\end{figure*}

% \subsection{Data in VSTAR}

The VSTAR is collected from 395 TV series (8,159 episodes, 185K 90-second clips) with carefully cleaned dialogues and metadata information. The collection and data cleaning details are as follow: 

\paragraph{Data Source} We carefully selected and purchased video sources from the copies of Blueray disks and online Reddit Open Directories. 
% In comparison to movies, TV series are closer to our life and contain less artistic expressions or monologues. 
% We carefully selected the TV series in order to ensure the quality of our dataset. 
Specifically, we filtered out animation and documentary series because the former differ a lot in terms of visual input styles and are unrealistic, while the latter mostly contain monologues. To ensure the quality of selected series, we selected the TV shows rated by over 1,000 IMDb users. The ultimate genre distribution is shown in \Cref{fig:genres}. As seen, there are 19 genres covering almost all common genres of TV series. For research purposes, we segmented each TV episode into 90-second video clips. In the end, we got 185K multi-modal dialogue clips.

\paragraph{Metadata} We crawled metadata for each episode from IMDb\footnote{\url{ https://www.imdb.com}} as complementary information. Each TV episode is paired with genres, keywords, and storylines. Compared to daily chat, the conversations in TV series are much longer and contain richer background knowledge. Therefore the metadata will be helpful for further dialogue understanding. We show the wordcloud of the story-lines in \Cref{fig:wordcloud}. It is interesting to observe that most salient words are relevant with ``work and life''. 

% \Cref{tab:VSTAR} shows comparisons between VSTAR and other video-grounded dialogue datasets.

\subsection{Annotation in VSTAR}

% \begin{figure}
%     \centering
%     \includegraphics[width=\linewidth]{figs/genre_epi.pdf}
%     % \begin{subfigure}{0.8\linewidth}
%     % \centering
%     %     \includegraphics[width=\linewidth]{figs/scene_len.pdf}
%     %     \caption{\# of session segments}
%     %     \label{fig:dist_scene}
%     % \end{subfigure}
%     % \hfill
%     % \begin{subfigure}{0.8\linewidth}
%     %     \centering
%     %     \includegraphics[width=\linewidth]{figs/session_len.pdf}
%     %     \caption{\# of topic segments}
%     %     \label{fig:dist_topic}
%     %  \end{subfigure}
%     \caption{Distribution of VSTAR Genres.}
%     \label{fig:genres}
% \end{figure}

\paragraph{Video Scene Boundary} A scene, according to the previous definition~\cite{Rasheed2003SceneDI,Huang2020MovieNetAH,Rao2020ALA}, is a plot-based semantic unit in which a certain activity occurs among a specific group of individuals. Recently popular methods use off-the-shelf shot segmentation tools first and  then determine whether the shot boundaries are scene boundaries. Considering the detection error of the shot segmentation tools, we did not take previous shot-based methods to annotate scene boundaries. Instead, we segment the TV episode into short videos by subtitle timeline. Specifically, each short video is paired with a dialogue turn. Annotators are then asked to look through these short videos with subtitles, and find if the short video is the start of a dialogue scene. With the help of multi-modal information, the dialogue scene boundaries are clearer. Thus the annotation procedure is more efficient. In order to keep consistency with previous work, we modified the annotated boundaries as the end of a dialogue scene. Finally we got 265k dialogue scene segments with 1.4 scene boundaries in each dialogue clip on average. Comparisons between VSTAR and other datasets for video scene segmentation are shown in \Cref{tab:scene}. We demonstrate that VSTAR is significantly larger than existing datasets. In addition, we show the distributions of the number of dialogue scenes in a TV episode in \Cref{fig:session_segments}. The number of scene segments in a TV episode is mostly between 10-60. To the best of our knowledge, VSTAR is the first dataset whose scene  boundaries are labeled with the help of multimodal information. 

\begin{table}[t!]
\centering
\resizebox{0.9\linewidth}{!}{
\begin{tabular}{lcccccc}
\thickhline
\#                & Scene   & Video     & Source    \\
\thickhline
\# OVSD           & 300   & 21   & MiniFilm \\
\hline
\# BBC            & 670   & 11   & Documentary \\ 
\hline
\# MovieScenes    & 21k   & 150      & Movie  \\ 
\hline
\# MovieNet       & 42k   & 318      & Movie  \\ 
\hline
\# VSTAR~(Ours)           & 265k  & 8159      & TV Episode  \\ 
\hline
\thickhline
\end{tabular}
}
\caption{Comparisons of dialogue scene annotation in VSTAR.}
\label{tab:scene}
\end{table}

\begin{table}[t]
\label{tab:topic}
\centering
\resizebox{\linewidth}{!}{
\begin{tabular}{lcccccc}
\thickhline
                & sentence   & sent/seg     & Language    \\
\thickhline
 DiaSeg\_711     & 19K   & 5.6   & English \\
\hline
 Doc2Dial       & 19K   & 3.5    & English \\ 
\hline
 ZYS            & 12K   & 6.4      & Chinese  \\ 
\hline
 VSTAR           & 4.6M  & 9.3      & English  \\ 
\hline
\thickhline
\end{tabular}
}
\caption{Comparisons of  dialogue topic annotations.}
\label{tab:topic}
\end{table}

% 话题切换时，我们把视频当作辅助信息来判断当前轮对话是不是话题中的最后一轮
\paragraph{Dialogue Topic Boundary} We perform dialogue topic boundary annotation and dialogue scene boundary annotation at the same time. Specifically, we take the video as auxiliary information to determine whether a dialogue turn is the end of a dialogue topic. In total, we obtain 499k dialogue topic segments with 2.7 topic boundaries in each dialogue clip on average. Each scene segment contains 1.88 topic segments. Comparisons between VSTAR and other datasets for dialogue topic segmentation are shown in \Cref{tab:topic}. VSTAR is $200\times$ larger than previous datasets in scale. And the dialogue topic length is longer than current datasets, which makes the dialogue topic segmentation task more challenging. As shown in \Cref{fig:topic_segments}, the number of dialogue topics in a TV episode varies from less than 10 to more than 160, which demonstrates the diversity of VSTAR.

\paragraph{Annotation Process}
We recruited 30 highly-educated students (undergraduates and above) with high English proficiency for the annotation. Each student is assigned with 4 groups of dialogues, each of which includes 40K continuous dialogue turns. For each dialogue group, we randomly sampled 5\% data and checked them manually. If the error rate is more than 4\%, we asked the annotator to re-annotate the whole dialogue turn sequence. We repeated this validation procedure three times. In the end, 4\% data did not meet our requirements and were all dropped off. During the annotation procedure, the salary for annotating each utterance is determined by the average time of annotation and local labor compensation standard.

\section{Benchmarks and Models}\label{sec:benchmark}

We set three benchmarks based on VSTAR for video-grounded dialogue understanding and generation. In Sec.\ref{sec:formu}, we first introduce the task formulation along with their evaluation metrics. Then in Sec.\ref{sec:model} we present our proposed transformer-based video dialogue frameworks to benchmark our tasks.

% the details of our sliding window-based segmentation model~(SWST). Lastly, we briefly illustrate the architecture of our autoregressive generative model~(AWDT) which utilizes the segmentation information to generate better responses.
% video scene segmentation ref: BASSL
% dialogue topic segmentation ref: improving-
% dialogue generation ref: openvidial & rlm

\subsection{Task formulation}
\label{sec:formu}
VSTAR consists of a set of video-grounded dialogue clips $(U, V) \in \mathcal{D}$, where $U=\{u_1, \ldots, u_N\}$ serves as a dialogue clip with $u_i$ denoting the $i$-th dialogue utterance, $V=\{v_1,\ldots,v_N\}$ signifies the corresponding video clip with $v_i$ paired with dialogue turn $u_i$, $N$ refers to the number of dialogue turns. More precisely, $v_i$ can be separated into a sequence of RGB image frames $\{z_{i,1}, \ldots, z_{i,K}\}$, where $z_{i,k}$ is a RGB tensor of frame $k$, $K$ is the number of frames in $v_i$.
% Note that each frame $z_j$ is a tensor with size of $(C, H, W)$ where $C$, $H$ and $W$ are RGB channels, the height and the width respectively. 

\paragraph{Video-grounded dialogue scene segmentation}
% A dialogue scene segment is a series of continuous short videos paired with dialogue turns. 
% We use $\{y^v_1, \ldots, y^v_N\}$ to denote dialogue scene segmentation labels, where $y^v_i \in \{0, 1\}$ indicates if $v_i$ is the last short video of a dialogue scene. 
A dialogue scene segment is a series of video-grounded dialogue pieces that sharing the same visual scene context.
We thereby formulate the dialogue scene segmentation task as a binary classification problem: given a clip $(U, V)=\{(u_i, v_i),i=1...K\}$, the model is asked to predict $s_i \in \{0, 1\}$ indicating the dialogue scene boundary. We take three commonly used metrics for evaluation:
\begin{itemize}[leftmargin=*,noitemsep, topsep=0pt]
    \item \textbf{AP}. We compute Average Precision~(AP) of $s_i=1$ for each video piece $v_i$.
    \item \textbf{mIoU}. Following \citet{Huang2020MovieNetAH}, we use mIoU to measure the averaged intersection-over-union~(IoU) between predicted dialogue scene segments and their closest ground truth dialogue scene segments.
    \item \textbf{micro-F1}. Inspired by \citet{Mun2022BoundaryawareSL}, we use micro-F1 as an additional evaluation metric to compare algorithms.
\end{itemize}

\paragraph{Video-grounded dialogue topic segmentation}
Similar to scene segmentation, we formulate dialogue topic segmentation as a turn-level classification problem. 
% $\{y^u_1, \ldots, y^u_N\}$ are dialogue topic segmentation labels, where $y_i^u \in \{0, 1\}$ indicates if $u_i$ is the last dialogue turn of a dialogue topic. 
Concretely, given a video-grounded dialogue clip $(U, V)$, we need to predict if the $i$-th dialogue utterance is the end of a dialogue topic. Following~\citet{Xing2021ImprovingUD}, we apply three standard metrics to evaluate the performance of the proposed segmentation model in this benchmark:
\begin{itemize}[leftmargin=*,noitemsep, topsep=0pt]
    \item \textbf{$P_k$ error score}~\cite{Beeferman2004StatisticalMF}. $P_k$ is a penalty computed via a sliding window of length $k$.
    \item \textbf{WinDiff}~\cite{Pevzner2002ACA}. WinDiff is calculated based on the intersection between reference segments and predicted segments within a moving window.
    \item \textbf{macro-F1}. We utilize the F1 score to make a balanced comparison of precision and recall.
\end{itemize}

\paragraph{Video-grounded dialogue response generation}
For each dialogue clip $(U,V)$, we set the first $N-1$ dialogue turns $\{u_1, \ldots, u_{N-1}\}$ as the dialogue context $C$ and the last dialogue turn $u_N$ as the gold reply $r$. We choose four commonly used reference-based metrics: BLEU~\cite{papineni2002bleu}, ROUGE~\cite{lin2004rouge}, METEOR~\cite{lavie2007meteor} and CIDEr~\cite{vedantam2015cider}.

\begin{figure*}[t!]
    \centering
    \includegraphics[width=\linewidth]{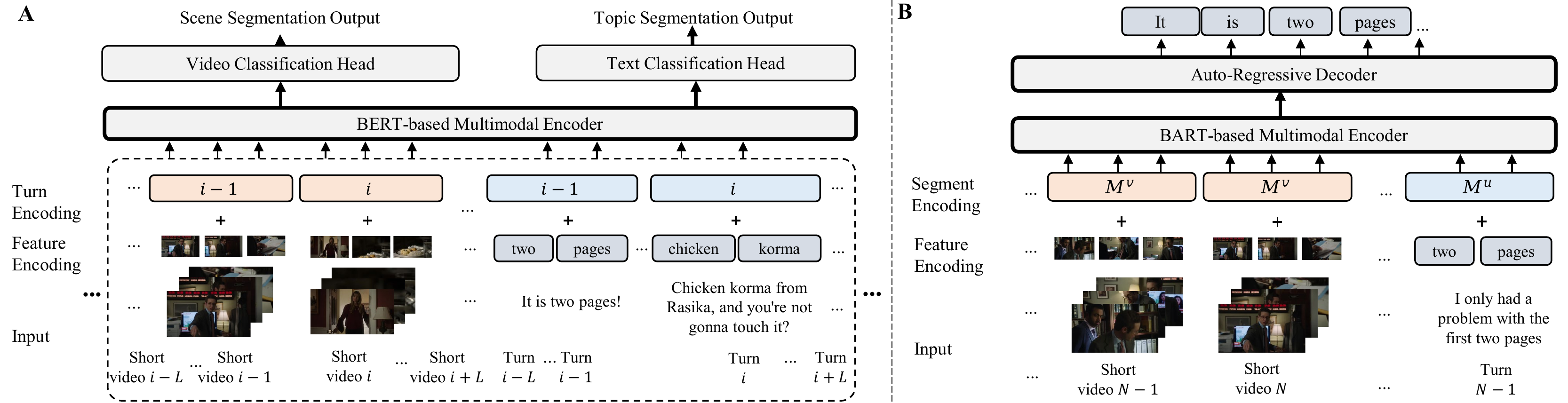}
    \caption{\textbf{Transformer-based model architecture.} (A) Sliding-window based segmentation transformer~(SWST) for scene and topic boundaries identification. The dashed rectangle indicates the current sliding window for turn $i$. 
    % For each turn, the encoder takes in  
    (B) Autoregressive video-grounded dialogue transformer~(AVDT) for dialogue response generation.}
    \label{fig:VSTAR_model}
\end{figure*}

\subsection{Transformer-based Video Dialogue Model}
\label{sec:model}
In this section, we propose a transformer-based discriminative model, namely SWST, to benchmark two segmentation tasks. For the response generation task, we develop a transformer-based generative model following encoder-decoder framework~(AVDT); \Cref{fig:VSTAR_model} depicts our overall architecture.

\paragraph{Sliding-window-based Segmentation Transformer~(SWST)}
 Inspired by currently popular works~\cite{Rao2020ALA,Chen2021ShotCS,Mun2022BoundaryawareSL} in video scene segmentation, we adopt the sliding window scheme to learn the contextual representation of a  dialogue scene. Specifically,
%  the window contains a sequence of $2L+1$ video-grounded dialogue pieces centered at the $(v_i,u_i)$, where $L$ is the number of neighbor pieces before and after $(v_i,u_i)$. Then 
the window can be denoted as a pair of short video sequence $W^v_i=\{v_{i-L},\ldots,v_i,\ldots,v_{i+L}\}$, and the corresponding dialogue turn sequence $W^u_i=\{u_{i-L}, \ldots, u_i, \ldots, u_{i+L}\}$, with $(v_i, u_i)$ as the center of the window, $K$ as the number of neighbor pieces before and after the center. Our goal is to train a model by maximizing the expected log-likelihood:
\begin{equation}
    \theta^* = \argmax_\theta \mathbb{E} \left[\log p_{\theta}(s_i|W^v_i, W^u_i) \right]
\end{equation}

\Cref{fig:VSTAR_model}A depicts the architecture of our Transformer-based video-grounded dialogue scene segmentation model. For visual feature $W^v_i$, we follow \citet{Lei2021LessIM} to randomly sample $M_l$ frames for $i$-th video piece $v_i$ instead of using the full-length short video. Then, we utilize ResNet-50~\cite{He2016DeepRL} pretrained on ImageNet dataset~\cite{Deng2009ImageNetAL} to extract 1,000 dim visual features for each frame. For dialogue feature $W^u_i$, we utilize the same tokenizer and embedding matrix as in BERT to obtain its initialization. We concatenate the sparse-sampled short video sequence $W^v_i$ and the dialogue turn sequence $W^u_i$ as the model input. The scene segmentation model leverages the same architecture and parameters in \textsc{BERT}\footnote{\url{https://huggingface.co/bert-base-uncased}}~\cite{Devlin2019BERTPO} as its initialization:
\begin{equation}
    e^v_i = f_{scene}([W^v_i;W^u_i]),
\end{equation}
where $f_{scene}: \mathbb{R}^{(4L+2) \times D_s} \rightarrow \mathbb{R}^{(4L+2) \times D_e}$ represents the \textsc{BERT}-based contextual relation network, $D_s$ and $D_e$ denote dimensions of the input and output features, $e^v_i = \{e^v_{i-L}, \ldots, e^v_i, \ldots, e^v_{i+L}\}$ represents the output feature sequence. After that, we apply a dialogue scene boundary detection head $h_S$ to predict the result for the contextualized representation. We then use cross-entropy loss to optimize the contextual relation network $f_{scene}$ and the dialogue scene boundary detection head $h_S$. In the test procedure, we binarize the prediction score with a threshold $\tau=0.5$ to get the result.

Similarly, we adopt the contextual relation network that has the same structure with $f_{scene}$ to encode the multi-modal inputs. And we use a linear layer as the dialogue scene boundary detection head, which is optimized by the ground truth dialogue topic labels.

\paragraph{Autoregressive Video-grounded Dialogue Transformer~(AVDT)}
Given a new dialogue context $C$ associated with a video clip $V$, our goal is to learn a generative model $p(r|V,C;\theta)$ from dialogue $\mathcal{D}$. \Cref{fig:VSTAR_model}(B) illustrates the architecture of our autoregressive generative model. The model is composed of a BART-based multi-layer Transformer~\cite{Lewis2020BARTDS} and an autoregressive decoder. We concatenate dialogue context $C = \{u_i\}_{i=1}^{N-1}$ and video clip $V=\{v_i\}_{i=1}^{N}$ as the encoder inputs. Considering the computing complexity, we sample one frame from each short video. Same as in the segmentation model, we use ResNet-50 to extract features of the sampled frames. We signify the dialogue scene segment sequence as $\{1,\ldots,M^v\}$ and the dialogue topic segment sequence as $\{1,\ldots,M^u\}$, where $M^v$ and $M^u$ are the number of scene segments in $V$ and topic segments in $C$, respectively. Then, we add these segment tokens to inputs to learn a scene \& topic aware context representation.

We adopt two other image embedding backbones to investigate the impact of the frame representation in our model: 
% Except for the CNN-based grid feature extractor, there are two other main-stream models to represent image feature:
OD-based region feature extractor and ViT-based patch feature extractor. Specifically, we use a Faster R-CNN~\cite{Ren2015FasterRT} trained on Visual Genome~\cite{Krishna2016VisualGC}  to extract the OD-based Region feature embedding. Each region feature is a 2048-d \ac{roi}. Following ViT~\cite{Dosovitskiy2021AnII}, we reshape the frame $Z_j \in \mathbb{R}^{(C\times H \times W)}$ into a sequence of flattened 2D patches $Z^p_j \in \mathbb{R}^{N \times (P^2\times C)}$, where $P$ is the size of each image patch and $N=H\times W/P^2$ is the number of patches.

\section{Experiments}\label{sec:exp}

We split the whole dataset into Train, Val, and Test with the ratio 17:1:1 on utterance level. For each task, we demonstrate the main results compared with baseline methods along with ablation studies on each module. Detailed information about baseline methods and implementation details are introduced in \Cref{sec:rel_work_ext} and \ref{sec:impl_details}.

% \subsection{Implementation Details}
% The whole dataset is split into Train, Val, and Test with the ratio 17:1:1 on utterance level. We use the pre-trained BERT$_{\rm base}$ model to initialize the parameters of our sliding-window based discriminative model. Besides, we choose the number of neighbor shot videos and dialogue turns $L=4$ and the sampled frame numbers $M_l=1$ after balancing the model performance and training  efficiency. We initialize the parameters of our autoregressive generative model by pre-trained BART$_{\rm base}$ model.

\subsection{Video Scene Segmentation}

We choose two popular methods as our baselines, \textbf{ShotCol}~\cite{Chen2021ShotCS} and \textbf{Bassal}~\cite{Mun2022BoundaryawareSL}. In practice, we did not extract the shot boundaries with external tools but used short video pieces as video units. We additionally implement the \textbf{Random} method by setting the ratio of scene segmentation label/non-segmentation label the same as in the test set. The overall results are shown in \Cref{tab:exp-scene}.  Compared with \textbf{ShotCol} and \textbf{Bassal} which focused on learning better shot-level representations based on ResNet-50, our method using offline-extracted video features can achieve similar performance with a difference within 2 points on F1 score. Concretely, our method performs better than ShotCol~\cite{Chen2021ShotCS} but worse than Bassal~\cite{Mun2022BoundaryawareSL}. This phenomenon shows that a better video encoder does help the model to distinguish the scene boundary. However, training a video encoder is really a time-consuming procedure. 

\textit{Ablation Studies} We ablated our method by adopting different lengths of sampled short videos. We find our method of taking 3 frames to represent the short video outperforms the 1-frame version. The frequency of change prediction between 1 sampled frame and 3 sampled frames is $0.137$. We believe the length of frames is an essential part not to be ignored for our transformer-based model. We also compare models with uni-modal inputs against ones with multi-modal inputs. The results demonstrate that with the help of text input, our method improves from 0.481 on mIoU to 0.536~($11.4\%$ relatively), 
from 0.474 on AP to 0.543~($14.6\%$ relatively), 
from 0.430 on F1 to 0.503~($17.0\%$ relatively). We are delighted to find the text information is very helpful to video scene segmentation.

\begin{table}[t!]
\centering
\resizebox{\linewidth}{!}{
\begin{tabular}{lccccc}
\thickhline
\multicolumn{1}{l}{Model}           & \multicolumn{1}{c}{mIoU~$\uparrow$} & \multicolumn{1}{c}{AP~$\uparrow$} & \multicolumn{1}{c}{F1~$\uparrow$}  \\ \hline

\multicolumn{1}{l}{Random}    & \multicolumn{1}{c}{0.251}     & \multicolumn{1}{c}{0.060}     & \multicolumn{1}{c}{0.075}        \\ 
% \hline
\multicolumn{1}{l}{ShotCol}             & \multicolumn{1}{c}{0.427}     & \multicolumn{1}{c}{0.412}     & \multicolumn{1}{c}{0.365}     \\ 
% \hline
\multicolumn{1}{l}{Bassal} & \multicolumn{1}{c}{0.466}     & \multicolumn{1}{c}{0.442}     & \multicolumn{1}{c}{0.401}    \\ 
% \hline

\multicolumn{1}{l}{SWST \textit{text-only}}             & \multicolumn{1}{c}{0.453}     & \multicolumn{1}{c}{0.351}     & \multicolumn{1}{c}{0.380}         \\ 

\multicolumn{1}{l}{SWST \textit{video-only~(1 frame)}}             & \multicolumn{1}{c}{0.448}     & \multicolumn{1}{c}{0.419}     & \multicolumn{1}{c}{0.385}         \\ 
% \hline
\multicolumn{1}{l}{SWST \textit{video-only~(3 frames)}}             & \multicolumn{1}{c}{0.481}     & \multicolumn{1}{c}{0.474}     & \multicolumn{1}{c}{0.430}         \\ 
% \hline
\multicolumn{1}{l}{SWST~\textit{(3 frames)}}        & \multicolumn{1}{c}{\textbf{0.536}}     & \multicolumn{1}{c}{\textbf{0.543}}     & \multicolumn{1}{c}{\textbf{0.503}}      \\  \thickhline
\end{tabular}
}
\caption{Results of dialogue scene segmentation task.}
\label{tab:exp-scene}
\end{table}

\begin{table}[t!]
\centering
\resizebox{0.9\linewidth}{!}{
\begin{tabular}{lccccc}
\thickhline
\multicolumn{1}{l}{Model}           & \multicolumn{1}{c}{WinDif~$\downarrow$} & \multicolumn{1}{c}{$P_k\downarrow$} & \multicolumn{1}{c}{F1~$\uparrow$}  \\ \hline

\multicolumn{1}{l}{Random}    & \multicolumn{1}{c}{0.765}     & \multicolumn{1}{c}{0.603}     & \multicolumn{1}{c}{0.370}        \\ 
% \hline
\multicolumn{1}{l}{TextTiling}  & \multicolumn{1}{c}{0.636}     & \multicolumn{1}{c}{0.581}     & \multicolumn{1}{c}{0.480}     \\ 
% \hline
\multicolumn{1}{l}{BERT+Greedy} & \multicolumn{1}{c}{0.615}     & \multicolumn{1}{c}{0.565}     & \multicolumn{1}{c}{0.486}    \\ 
% \hline
\multicolumn{1}{l}{BERT+CS}             & \multicolumn{1}{c}{0.541}     & \multicolumn{1}{c}{0.512}     & \multicolumn{1}{c}{0.527}         \\ 
% \hline
\multicolumn{1}{l}{BERT+CS SP}             & \multicolumn{1}{c}{0.531}     & \multicolumn{1}{c}{0.422}     & \multicolumn{1}{c}{0.610}         \\ 
% \hline
\multicolumn{1}{l}{SWST \textit{text-only}}        & \multicolumn{1}{c}{0.374}     & \multicolumn{1}{c}{0.326}     & \multicolumn{1}{c}{0.644}      \\ 
% \hline
\multicolumn{1}{l}{SWST \textit{full}}            & \multicolumn{1}{c}{\textbf{0.336}}     & \multicolumn{1}{c}{\textbf{0.281}}     & \multicolumn{1}{c}{\textbf{0.690}}       \\  \thickhline
\end{tabular}
}
\caption{Results of the dialogue topic segmentation task.}
\label{tab:exp-topic}
\end{table}

\begin{table*}[t!]
\centering
\resizebox{\linewidth}{!}{
\begin{tabular}{lccccccc}
\thickhline
\multicolumn{1}{l}{Model}           & \multicolumn{1}{c}{BLEU1} & \multicolumn{1}{c}{BLEU2} & \multicolumn{1}{c}{BLEU3} & \multicolumn{1}{c}{BLEU4} & \multicolumn{1}{c}{METEOR} & \multicolumn{1}{c}{ROUGE-L} & CIDEr \\ \hline

\multicolumn{1}{l}{OpenViDial coarse}    & \multicolumn{1}{c}{0.075}     & \multicolumn{1}{c}{0.026}     & \multicolumn{1}{c}{0.013}     & \multicolumn{1}{c}{0.006}     & \multicolumn{1}{c}{0.035}      & \multicolumn{1}{c}{0.063}       & 0.066     \\ 
% \hline
\multicolumn{1}{l}{RLM}    & \multicolumn{1}{c}{0.072}     & \multicolumn{1}{c}{0.032}     & \multicolumn{1}{c}{0.017}     & \multicolumn{1}{c}{0.010}     & \multicolumn{1}{c}{0.032}      & \multicolumn{1}{c}{0.061}       & 0.079     \\ 
% \hline
\multicolumn{1}{l}{AVDT} & \multicolumn{1}{c}{\textbf{0.089}}     & \multicolumn{1}{c}{\textbf{0.040}}     & \multicolumn{1}{c}{\textbf{0.022}}     & \multicolumn{1}{c}{\textbf{0.014}}     & \multicolumn{1}{c}{\textbf{0.041}}      & \multicolumn{1}{c}{\textbf{0.082}}       &   \textbf{0.145}   \\ 
\hline
\multicolumn{1}{l}{AVDT \textit{w/o seg query-only}} & \multicolumn{1}{c}{0.082}     & \multicolumn{1}{c}{0.035}     & \multicolumn{1}{c}{0.019}     & \multicolumn{1}{c}{0.011}     & \multicolumn{1}{c}{0.037}      & \multicolumn{1}{c}{0.073} & 0.119     \\ 
% \hline
\multicolumn{1}{l}{AVDT \textit{w/o seg text-only}} & \multicolumn{1}{c}{0.085}     & \multicolumn{1}{c}{0.037}     & \multicolumn{1}{c}{0.021}     & \multicolumn{1}{c}{0.013}     & \multicolumn{1}{c}{0.037}      & \multicolumn{1}{c}{0.075} & 0.126     \\ 

% \hline
\multicolumn{1}{l}{AVDT \textit{w/o seg (1 frame)}} & \multicolumn{1}{c}{0.087}     & \multicolumn{1}{c}{0.039}     & \multicolumn{1}{c}{0.021}     & \multicolumn{1}{c}{0.013}     & \multicolumn{1}{c}{0.039}      & \multicolumn{1}{c}{0.077}       &   0.126   \\

% \hline
\multicolumn{1}{l}{AVDT \textit{w/o seg (3 frames})} & \multicolumn{1}{c}{\textbf{0.090}}     & \multicolumn{1}{c}{\textbf{0.040}}     & \multicolumn{1}{c}{\textbf{0.022}}     & \multicolumn{1}{c}{\textbf{0.014}}     & \multicolumn{1}{c}{\textbf{0.041}}      & \multicolumn{1}{c}{\textbf{0.081}}       &   \textbf{0.139}   \\ 

\hline
% \multicolumn{1}{l}{resnet embed + segment embed} & \multicolumn{1}{c}{0.089}     & \multicolumn{1}{c}{0.040}     & \multicolumn{1}{c}{0.022}     & \multicolumn{1}{c}{0.014}     & \multicolumn{1}{c}{0.041}      & \multicolumn{1}{c}{0.082}       &   0.145   \\ 
% \hline
\multicolumn{1}{l}{AVDT~\textit{Faster-RCNN}}             & \multicolumn{1}{c}{0.089}     & \multicolumn{1}{c}{0.040}     & \multicolumn{1}{c}{0.022}     & \multicolumn{1}{c}{0.014}     & \multicolumn{1}{c}{0.040}      & \multicolumn{1}{c}{0.080}       &  0.137    \\ 
% \hline
\multicolumn{1}{l}{AVDT \textit{ViT}}        & \multicolumn{1}{c}{\textbf{0.092}}     & \multicolumn{1}{c}{\textbf{0.041}}     & \multicolumn{1}{c}{\textbf{0.023}}     & \multicolumn{1}{c}{\textbf{0.014}}     & \multicolumn{1}{c}{\textbf{0.041}}      & \multicolumn{1}{c}{\textbf{0.082}}       &  \textbf{0.144}    \\ 
\thickhline
\end{tabular}
}
\caption{Results of the dialogue response generation task. AVDT denotes our Autoregressive Video-grounded Transformer. All evaluation metrics denote better generation performance with higher scores.}
\label{tab:exp-gen}
\end{table*}

\begin{figure*}[t!]
    \centering
    \resizebox{\linewidth}{!}{
    \setlength{\tabcolsep}{2pt}
    \begin{tabular}{p{1.2in}p{1.2in}p{1.2in}p{1.2in}p{1.2in}p{1.2in}}
        \includegraphics[width=\linewidth]{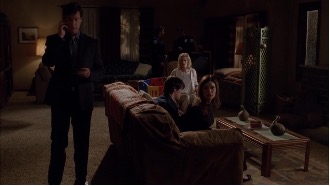} & 
        \includegraphics[width=\linewidth]{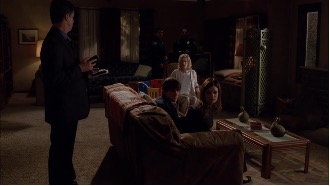} & 
        \includegraphics[width=\linewidth]{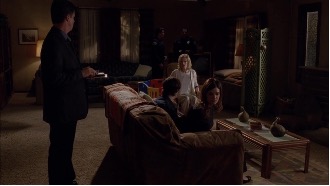} & 
        \includegraphics[width=\linewidth]{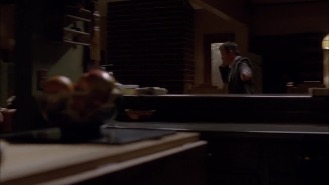} & 
        \multicolumn{2}{c}{\includegraphics[width=1.2in]{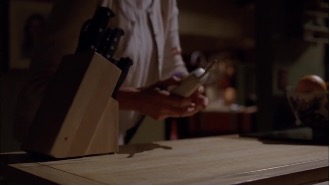}}
        
        \\ 
        Pick up. Pick up. & 
        That's him. That's my husband. &
        Start tracing the landline. &
        The number's 177-8987. -Skyler. &
         \parbox[t]{1.1in}{\textbf{REF:} I know you are there, so pick it up.}
        &
        \textbf{PRED:} I need you to pick up the phone. (ROUGE=0.44) 
\\
        \includegraphics[width=\linewidth]{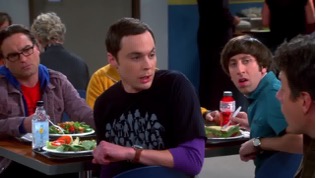} & 
        \includegraphics[width=\linewidth]{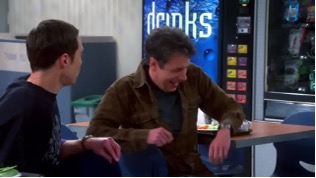} & 
        \includegraphics[width=\linewidth]{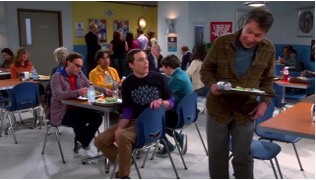} & 
        \includegraphics[width=\linewidth]{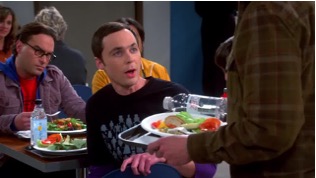} & 
        \multicolumn{2}{c}{\includegraphics[width=1.2in]{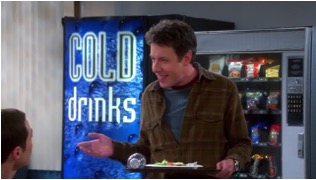}} \\ 
        Did they find evidence to support extra dimensions or supersymmetry? &
        No, but they did find evidence that you'll believe anything. &
        Why would you do that? &
        You're a string theorist as well &
        \parbox[t]{1.1in}{\textbf{REF:} Incorrect. I am a string pragmatist.} &
        \textbf{PRED:} I'm a string theorist. (ROUGE=0.22)
    \end{tabular}}
    \caption{Examples of human evaluation. Each image indicates a short video clip, \textbf{PRED} denotes our predicted response, \textbf{REF} denotes the reference human response.}
    \label{fig:human_case}
\end{figure*}

\subsection{Dialogue topic segmentation}
We compare several currently popular baselines: \textbf{TextTiling}~\cite{Hearst1997TextTS}, \textbf{GreedySeg}~\cite{Xu2021TopicAwareMD} and  \textbf{BERT+CS}~\cite{Xing2021ImprovingUD}. In addition, we train another \textbf{BERT+CS SP} model under the supervision of the ground truth label for comparison.  We implement \textbf{Random} algorithm following~\cite{Xing2021ImprovingUD}. The overall results are represented in \Cref{tab:exp-topic}. The \textbf{BERT+CS SP} model with supervision signals improves 0.527 on F1 to 0.610 ($15.7\%$ relatively). This result shows the importance of our dialogue topic boundary annotations. Compared with \textbf{BERT+CS SP} model, our sliding window-based approach SWST improves 0.610 on F1 to 0.644($5.6\%$ relatively).  To further investigate the validity of the visual information in dialogue topic segmentation, we add the video clip to the inputs. The result turns out that the visual information is important for the dialogue topic boundary detection where it leads to a performance  gain of $7.1\%$ in F1.

\subsection{Video-grounded Response Generation}
We choose two commonly used transformer-based models as our baselines: \textbf{CoarseVisual}~\cite{Wang2021ModelingTM} and \textbf{RLM}~\cite{li2020bridging}.  Visual features in these models are all extracted through ResNet-50 pre-trained on ImageNet dataset. 
The overall results are shown in \Cref{tab:exp-gen}. Results show that our model outperforms baseline methods across all metrics.
% Since open-domain dialogue generation is still a challenging task, the differences among the results of test methods are not significant as expected. 
% Our observations are as follows:

 \textit{Analysis \& Ablation studies} We further examined the performance of our model on different input settings~(rows 4-7) and made three fundamental observations: (i) When the input is only a query~(row 4), our model is slightly worse than the input is the full dialogue clip history~(row 5), which demonstrates that the dialogue history containing much noise is of limited help for response generation. (ii) Though our model using the text-only setting~(row 5) can reach comparable results with the standard setting~(row 6), the results improve across all metrics when we increase the number of input frames (rows 6-7). Such observation is similar to \citet{Huang2021WhatMM} --- the utility of multi-modality is positively correlated with the data scale.  (iii) Compared with the setting without segment embedding~(row 6), our method performs much better. Specifically, the Rouge-L score increases from 0.077 to 0.082 with an increment of $6.5\%$; the CIDEr increases from 0.126 to 0.145 with an increment of $15.1\%$. These improvements show that segment information is important for dialogue generation in our dataset. 

 Additionally, we investigate the contribution of different visual backbones. Specifically, we use different frame representations while keeping other parts of our model unaltered. The results (rows 8-9) show similar performance between these feature representations and the ViT-based patch features even perform slightly better than other offline-extracted features with high computational overload. This phenomenon validates our hypothesis that the current encoder-decoder model can not make full use of the visual information for the video-grounded dialogue generation task, which yields future investigation on video-grounded dialogue modeling.

\begin{table}[ht]
\label{tab:human_eval}
\centering
\resizebox{\linewidth}{!}{
\begin{tabular}{lcccccc}
\thickhline
                & Win   & Lose     & Tie & Kappa    \\
\thickhline
AVDT \vs OpenViDial     & 0.20   & 0.16   & 0.64 & 0.71 \\
% \hline
 AVDT \vs Human       & 0.08   & 0.71    & 0.21 & 0.73 \\ 
\hline
\thickhline
\end{tabular}
}
\caption{Human evaluation result}
\label{tab:human_eval}
\end{table}
\textit{Human Evaluation} We follow \citet{sun-etal-2022-multimodal} to run human evaluation by comparing our generated responses with baseline methods. Specifically, we select 20 highly educated students with proficient English skills as evaluators and randomly sample 300 video-dialogue pairs with corresponding responses as test cases. To each evaluator, two responses from different models are presented, which are shuffled to hide the sources. The evaluators then judge which response is more coherent to the current dialogue scene and topic. The agreement among the annotators is measured by Fleiss' Kappa~\cite{Fleiss1971MeasuringNS}. \Cref{tab:human_eval} shows the human evaluation results and \Cref{fig:human_case} shows some qualitative comparison. We additionally add a comparison with human annotation as an upper bound.

% \textit{Case Study}

\section{Limitations and Ethics Concerns}
% 数据集中的对话是人写的和日常对话不一样

We point out potential limitations and ethical concerns of this work.

\noindent\textbf{Limitation: Data and Modeling}
 The dialogues in our dataset are made by playwright, which are slightly different from daily chat. Second, the automatic evaluation metrics for the response generation task can not perfectly reflect the interactiveness of dialogue system. Lastly, Our autoregressive generative model simply add the segment embedding to the inputs. Similar to the position encoding in transformer, our coarse method does not make good use of the segmentation, and lacks interpretability.

\noindent\textbf{Ethics: Copyright and Licensing}
The data source are publicly available TV series. Its collection and annotation procedure is designed for video-grouned dialogue understanding and generation purpose, and does not involve privacy issues. Following  LSMDC~\cite{Rohrbach2016MovieD} and MovieNet~\cite{Huang2020MovieNetAH}, we will polish an agreement and release TV shows content under very strict conditions but will open-source all the scrawling code, pretrained features and sampled images.

% \noindent\textbf{Potential Data Noise}
% Although we take principle measures to ensure the quality of the dataset, there might still be moderate noise our large-scale dataset.

\section{Conclusion}

In this paper, we introduce VSTAR, a scene and topic-aware video-grounded dialogue understanding and generation dataset. 
The main purpose of our dataset is to improve the situated multimodal semantic perception capability of the dialogue system, so that the dialogue system can generate the response that are both semantically and logically consistent with the dialogue scene in common situation. 
We introduce three challenging benchmarks on different aspects of video-grounded dialogue understanding and generation, \ie, discovering scene transition and topic transition on video-grounded dialogue, and getting proper response. Furthermore, we propose three new baselines for corresponding benchmarks. Experiment results shows the multi-modal information can benefit dialogue understanding. Besides, we find scene boundary and topic scene boundary contribute to generate more relevant and coherent responses.
By introducing VSTAR, we hope to shed light on future research towards building conversational agents that can comprehend complicated realistic multimodal signals.

\section*{Acknowledgements}
% This work is supported in part by National Key R\&D Program of China (2021ZD0150200).
This work is supported by the National Key Research and Development Program of China (No.2020AAA0106600).

\bibliography{custom}
\bibliographystyle{acl_natbib}

\appendix

\end{document}